\documentclass[letterpaper, 10 pt, conference]{ieeeconf}  

\IEEEoverridecommandlockouts                              

\usepackage{cite}
\usepackage[T1]{fontenc}
\usepackage[utf8]{inputenc}
\usepackage{graphicx}

\usepackage{amsmath}
\usepackage{amssymb}
\usepackage{booktabs}
\usepackage{lmodern}   
\usepackage{multirow}

\title{\LARGE \bf
TS-Mask VLA: 2D Temporal--Spatial Masking for Vision-Language-Action Model with Effective Bridging
}

\author{Shengzhuo Yang$^{1*}$, Ronghao Yu$^{1}$, Chuanjie Lv$^{1}$, Linpeng Peng$^{1}$,\\ Hang Yu$^{2}$, Jie Ren$^{1}$, Jiajun Lv$^{1}$ and Yong Liu$^{1\dagger}$
\thanks{*This work was supported by “Zhejiang Key Laboratory of Advanced Intelligent Warehousing and Logistics Equipment” (Grant No. 2024E10007)}
\thanks{$^{1}$The authors are with the Institute of Cyber-Systems and Control, Zhejiang University, Hangzhou, China.}%
\thanks{$^{2}$Hang Yu is with Tongji University, Shanghai, China.}%
\thanks{$^{*}$Equal contribution.}%
\thanks{$^{\dagger}$Corresponding author (Email: yongliu@iipc.zju.edu.cn).}%
}

\begin{document}

\maketitle
\thispagestyle{empty}
\pagestyle{empty}

\begin{abstract}
Vision--language--action (VLA) models aim to understand natural-language instructions and visual observations, and to generate and execute corresponding actions as embodied agents. Recently, autoregressive token-based action generation has driven the development of many representative VLA models. However, this paradigm often reduces action generation to next-token prediction, thereby lacking explicit modeling of the spatiotemporal structure of action sequences and the disentanglement between vision-language representations and actions, which can limit performance in long-horizon and complex scenarios. In this paper, we propose TS-Mask VLA, a vision--language--action framework for robot manipulation. TS-Mask VLA is built upon two key designs: (1) We propose a Discrete Diffusion Action Expert equipped with a Bridge Attention conditioning bridge, which enables multi-layer conditioning from the VLM and facilitates more accurate and stable action generation; and(2) We propose a temporal--spatial 2D masking strategy for discrete action tokens that strengthens the model's understanding of cross-time dependencies and inter-dimensional coupling, leading to more structurally consistent action sequences; and 
We conduct extensive experiments on simulation benchmarks and real-world tasks. 
On LIBERO, TS-Mask VLA achieves a 95.7\% average success rate with only 0.5B parameters, outperforming significantly larger models. 
On CALVIN, it attains the best average sequence length of 4.19 and strong long-horizon performance. 
Comprehensive analyses and ablations further validate the effectiveness of our design.
\end{abstract}

\section{INTRODUCTION}
In the past two years, rapid advances in multimodal large language models have made embodied robotic systems with general perception, understanding, and behavior a central research focus. Vision--Language--Action (VLA) models align visual perception, language understanding, and action generation within a unified framework, enabling instruction-driven robot manipulation. Most existing VLA build upon a pre-trained vision--language backbone with an action-generation head that maps observations and instructions to action sequences, following two dominant paradigms: (i) autoregressive transformers that sequentially predict action tokens\cite{liang2025discrete,wen2025llada,kim2024openvla}, and (ii) diffusion-based approaches that generate continuous trajectories via iterative denoising\cite{chi2025diffusion,shukor2025smolvla}. Despite this progress, VLA still face challenges in long-horizon tasks and complex environments, particularly in balancing structural unification and efficiency.

Current VLA methods predominantly follow two paradigms, both exhibiting inherent limitations.
The first unifies vision, language, and actions into a single token sequence and autoregressively generates actions\cite{kim2024openvla}, lacking explicit structural decoupling between representation learning and control policy modeling.
The second adopts continuous diffusion to iteratively generate trajectories\cite{reuss2024multimodal}, which can lead to instability over long horizons and makes it difficult to explicitly model temporal--spatial correlations in robotic actions.
Motivated by these observations, we argue that two aspects are crucial:
(i) bridging vision-language representations into a dedicated action expert to establish clearer conditional decoupling; and
(ii) discretizing actions to explicitly model their structured dependencies, enabling more robust generalization in complex environments.

To solve these problems, we propose TS-Mask VLA, a discrete VLA framework that integrates a 2D temporal--spatial masking strategy with effective vision--language conditioning to explicitly model the structured dependencies of action sequences and generate robust actions. Specifically, we first extract visual features from both the third-view camera and the wrist camera, and feed them, together with the language instruction, into a lightweight VLM backbone. Inspired by VLA-Adapter\cite{Vla-adapter}, we leverage intermediate features from all VLM layers and inject them into a dedicated discrete diffusion action expert in a layer-aligned manner. Within this expert, we adopt a Bridge Attention mechanism to effectively transfer vision--language conditional information into the action space, thereby directly guiding action generation.

Next, inspired by discrete diffusion models\cite{liang2025discrete}, we discretize continuous actions into tokens and organize them into a two-dimensional discrete action structure defined over time steps and action dimensions. 
Based on this representation, we introduce a Temporal--Spatial 2D Mask strategy. 
Unlike conventional 1D random masking\cite{javed2024intermask} that disrupts sequences without structural awareness, our method applies structured perturbations along both the temporal axis and the action-dimension axis. 
This explicitly strengthens the modeling of cross-time dependencies and inter-dimensional coupling, while encouraging the model to focus on action-relevant regions and key time steps. 
Compared with continuous diffusion formulations that treat trajectories as holistic real-valued signals, such structured discrete modeling provides a clearer inductive bias over the intrinsic time and action-dimension structure, thereby enhancing the modeling of spatiotemporal correlations in action sequences.

During training, we adopt masked denoising learning by randomly masking a subset of positions on the 2D action grid and training the model to recover the masked action tokens under vision--language conditioning and proprioceptive states, thereby learning to robustly complete actions from incomplete contexts; During inference, we start from a fully masked 2D action grid and perform multiple rounds of parallel refinement: at each iteration, the model predicts candidate action tokens and applies a ReMask procedure that keeps high-confidence predictions while re-masking low-confidence ones, progressively refining the action sequence in a coarse-to-fine manner as the masking ratio decreases.

We systematically evaluate TS-Mask VLA on two standard benchmarks, LIBERO\cite{liu2023libero} and CALVIN\cite{mees2022calvin}, and achieve strong performance on both. Despite using only a 0.5B model, our approach consistently outperforms several significantly larger VLA baselines, including OpenVLA\cite{kim2024openvla} and $\pi_0$\cite{black2024pi_0}, demonstrating a favorable performance. Real-world evaluations further validate its practical applicability and generalization in physical robotic settings. Overall, our contributions are threefold:

\begin{itemize}
    \item We propose a Discrete Diffusion Action Expert equipped with a Bridge Attention module, which injects hierarchically aligned multi-layer conditioning from the VLM to enable more effective action generation.
    \item We propose a temporal--spatial 2D masking strategy for discrete action tokens, allowing the model to better understand key temporal dependencies and inter-dimensional coupling within action sequences, and thus produce more structurally consistent actions.
    \item Experiments show that TS-Mask VLA achieves high success rates across diverse simulation benchmarks and real-world robotic tasks with a smaller model scale, demonstrating strong generalization and practical deployability.
\end{itemize}



\section{Related Work}
\noindent\textbf{Vision-Language-Action Models}\quad
Recent advances in large-scale Vision--Language Models (VLMs) have introduced a new paradigm for robotic control. By extending visual and linguistic representations to action prediction, researchers have developed the Vision--Language--Action (VLA) framework. Representative works such as RT-1~\cite{Rt-1} and RT-2~\cite{Rt-2} discretize continuous control signals into tokens, enabling compatibility with text-based VLM architectures, and OpenVLA~\cite{kim2024openvla}, as the first open-source VLA model, further accelerates progress in this direction. Building upon this foundation, subsequent studies improve action modeling from multiple perspectives; for example, $\pi_0$~\cite{black2024pi_0} leverages flow-matching with large-scale multi-task datasets to achieve strong control performance. In addition, VLA-Adapter~\cite{Vla-adapter} employs parameter-efficient fine-tuning to adapt pretrained VLMs to robotic control tasks while reducing training cost. Despite these advances, most existing VLA frameworks are built upon autoregressive VLMs, where action generation relies on sequential prediction\cite{kim2024openvla, Rt-1, Rt-2}. In contrast, discrete-diffusion-style vision--language--action modeling has received relatively less attention in the VLA literature. We therefore investigate how to build a discrete diffusion-based framework on top of pretrained diffusion-style vision--language models to achieve more stable and structurally coherent action generation.

\noindent\textbf{Discrete Diffusion Models.}\quad
Discrete diffusion extends diffusion-based generation to categorical token spaces by formulating the forward process as a Markov chain that progressively corrupts discrete tokens~\cite{austin2021structured,hoogeboom2021argmax}.
Unlike continuous diffusion in Euclidean space, discrete diffusion operates directly over categorical distributions, making it well-suited to text and tokenized image modeling~\cite{gu2022vector,li2022diffusion}.
Masked discrete diffusion models have shown competitive performance against autoregressive approaches in both image and language generation, while enabling parallel denoising and controllable inference~\cite{chang2022maskgit}.
Recent works further scale this paradigm to large language and multimodal models~\cite{li2022diffusion,nie2025large}.
However, its application to robotic action modeling remains relatively limited.
Compared to text and images, action sequences exhibit structured temporal dependencies and control constraints, posing unique challenges for diffusion-based modeling.
In this work, we extend discrete diffusion to tokenized action chunks within a discrete diffusion action expert, enabling structured and parallel denoising for more stable action generation and advancing unified modeling across vision, language, and action.

\section{METHOD}


\subsection{Overview} 
\label{sec:overview}
\begin{figure*}[t]
    \centering
    \includegraphics[width=\textwidth]{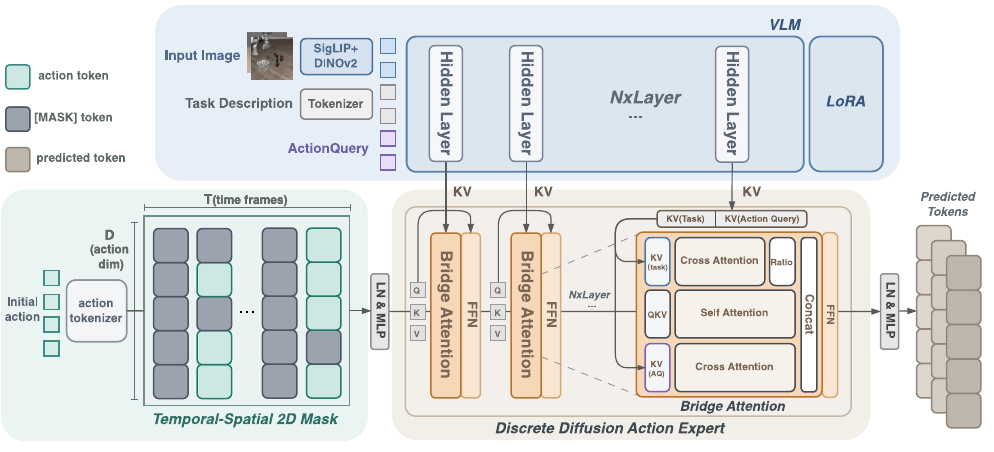}
    \caption{\textbf{Overview of TS-Mask VLA.}
    The input observation image and task description are encoded by a visual backbone and a text tokenizer, and together with Action Query fed into an $N$-layer VLM to produce hidden states, where we adopt Qwen2.5-0.5B as the backbone. The VLM hidden states are provided as key/value to a Discrete Diffusion Action Expert composed of $N$ stacked blocks of Bridge Attention and FFN. On the action side, initial action is discretized by an action tokenizer and organized with a temporal--spatial 2D mask over time frames and action dimensions, where \texttt{[MASK]} tokens indicate masked positions and predicted tokens are progressively denoised. The action expert fuses VLM context with the action-token sequence via bridge attention and outputs the final predicted action tokens. Color coding denotes action tokens, \texttt{[MASK]} tokens, and predicted tokens.
    }
    \label{fig:overview}
\end{figure*}

We propose TS-Mask VLA, whose overall architecture is illustrated in Fig.~\ref{fig:overview}. The framework consists of three core components: (1) a VLM backbone that encodes visual observations and text instructions to produce fused multimodal representations; (2) a Discrete Diffusion Action Expert that generates discrete action tokens under multimodal conditioning and progressively refines them via iterative denoising; and (3) a temporal--spatial masking module that applies 2D masking over discretized action tokens to explicitly capture temporal dependencies and inter-dimensional coupling. Sec.~\ref{sec:vlm} explains how multimodal information is injected from the VLM backbone, Sec.~\ref{sec:ddae} presents the Discrete Diffusion Action Expert, Sec.~\ref{sec:mask} introduces the temporal--spatial action masking strategy, and Sec.~\ref{sec:Training} details the training and inference procedures.

\begin{figure}[t]
    \centering
    \includegraphics[width=\columnwidth]{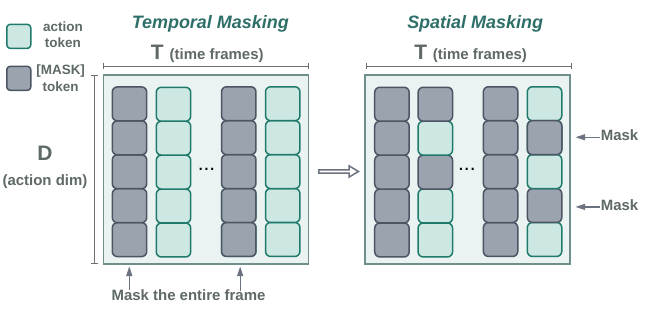}
    \caption{\textbf{Temporal--Spatial 2D Token Masking.}
    Given a 2D action-token map of size $T \times D$, we adopt a two-stage masking strategy. We first perform Temporal masking (left), where several time frames are randomly selected and all action tokens within the selected frames are masked. Then, Spatial masking (right) is applied to the remaining unmasked frames, where a subset of action tokens within each frame is randomly masked.}
    \label{fig:temporal_spatial_masking}
\end{figure} 
\subsection{Multimodal VLM encoding}
\label{sec:vlm}
For visual encoding, we employ the DINOv2\cite{oquab2023dinov2} and SigLIP\cite{zhai2023sigmoid} to extract features from the third-view image $I_t^{v}$ and the gripper image $I_t^{g}$, and project them into the language embedding space via a fused MLP projector. The instruction $\mathcal{L}_t$ is tokenized by the backbone tokenizer. We adopt Qwen2.5-0.5B\cite{hui2024qwen2} as the backbone of our VLM, achieving an efficient yet expressive multimodal representation.
At timestep $t$, the VLM takes multimodal inputs:
\begin{equation}
\bigl(I_t^{v},\, I_t^{g},\, \mathcal{L}_t,\, AQ_t \bigr),
\end{equation}
where $AQ_t$ is a learnable Action Query token appended to the input sequence to provide action-centric guidance. Instead of only using the final-layer output, we extract hidden states from all $L$ Transformer layers:
\begin{equation}
\mathbf{H}=\{H^{(l)}\}_{l=1}^{L},
\end{equation}
where $L$ denotes the number of layers and $H^{(l)}\in\mathbb{R}^{N\times d}$ is the token-wise hidden state matrix of the $l$-th layer, with $N$ tokens and hidden dimension $d$. These layer-wise features provide complementary cues ranging from fine-grained spatial details to high-level semantics.
We inject $\mathbf{H}$ into the Discrete Diffusion Action Expert via layer-aligned conditioning, enabling the diffusion policy to jointly leverage low-level perceptual signals and high-level instruction grounding during discrete action token generation. The corresponding Action Query(AQ) features are also used as conditioning signals, offering more action-centric multimodal guidance.

\begin{figure*}[t]
    \centering
    \includegraphics[width=\textwidth]{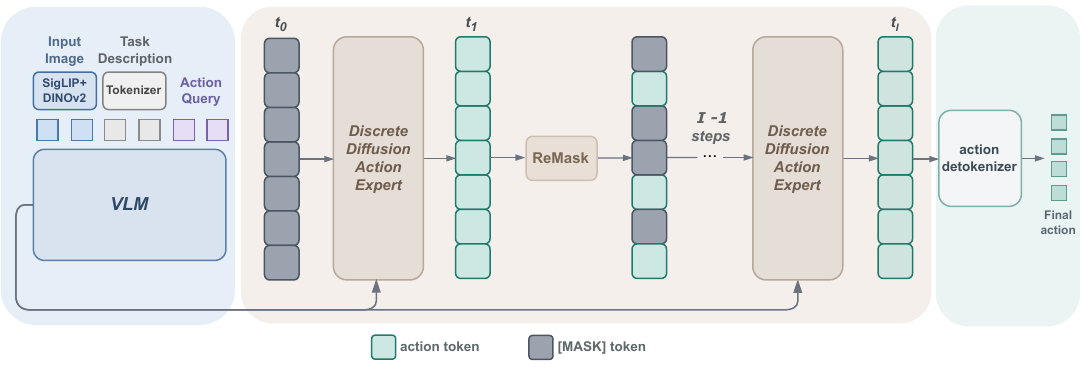}
    \caption{\textbf{Inference procedure of TS-Mask VLA.}
    Given an input image, task description, and Action Query, the VLM encodes multimodal context and conditions the Discrete Diffusion Action Expert. Starting from a fully masked action-token sequence at $t_0$, the model predicts token distributions and produces partially denoised tokens at $t_1$. A ReMask strategy is then applied to re-mask low-confidence tokens while preserving high-confidence predictions. This procedure is repeated for $I-1$ additional steps, progressively refining the action-token sequence from $t_0$ to $t_I$. Finally, the fully denoised discrete tokens are decoded by the action detokenizer to obtain the final continuous action. Colors indicate action tokens and \texttt{[MASK]} tokens.
    }
    \label{fig:inference}
\end{figure*}

\subsection{Discrete Diffusion Action Expert}
\label{sec:ddae}
We propose a discrete diffusion action expert module, which generates discrete action tokens conditioned on the multi-layer hidden states of VLM backbone. 
Given the layer-wise hidden layer vectors of the VLM backbone, where each layer contains visual features and AQ features, we construct an $L$-layer discrete diffusion action expert aligned with the corresponding layers of the backbone network, thereby enabling layer-wise feature injection.

\subsubsection{Bridge Attention Block}
\label{sec:bridge}
Inspired by VLA-Adapter, we adopt Bridge Attention block, which fuses three information
sources within one attention computation:
(i) Self tokens: derived from the current action representations to model
intra-action temporal dependencies;
(ii) AQ tokens: Action Query features extracted from the VLM hidden states, serving as action-oriented conditioning signals to guide the refinement of the action representations; and
(iii) Task tokens: visual--language features from the VLM hidden states,
providing external task constraints for policy representation.

Concretely, we form a shared query $Q$ from the current action features,
while deriving source-specific keys and values $\big(K_{\text{self}}, V_{\text{self}}\big)$,
$\big(K_{\text{AQ}}, V_{\text{AQ}}\big)$, and
$\big(K_{\text{task}}, V_{\text{task}}\big)$for each stream via
independent MLP projections. The attention logits from the three sources are
concatenated along the key dimension and jointly normalized:

\begin{equation}
\alpha=\mathrm{softmax}\!\left(
\frac{\left[
QK_{\text{self}}^{\top},\;
QK_{\text{AQ}}^{\top},\;
\tanh(g)\,QK_{\text{task}}^{\top}
\right]}{\sqrt{d_h}}
\right),
\end{equation}
where $\alpha$ denotes the normalized attention weights over the
concatenated key tokens from all streams, and $g$ is a learnable gating scalar that modulates the task branch, mitigating excessive task conditioning during early training.
The bridge attention output is then computed as
\begin{equation}
\mathrm{Z}_{bridge}
=
\alpha
\left[
V_{\text{self}},\;
V_{\text{AQ}},\;
V_{\text{task}}
\right],
\end{equation}
where $Z_{\text{bridge}}$ denotes the output of the bridge attention block, obtained by aggregating values from the self, AQ, and task streams under the cross-stream attention weights $\alpha$.
To preserve positional structures within each sequence, Rotary Position Embedding (RoPE) is applied independently to each key branch. Finally, the attention output is
linearly projected and added back via a residual connection, followed by a
feed-forward network to produce the updated representation at layer $l$.

\subsubsection{Uniform Quantized Action Tokenization}
\label{sec:Tokenization}
To discretize continuous actions, we adopt a \emph{uniform quantized} action-token representation. Each normalized scalar:
\begin{equation}
\hat{a}_i \in [-1,1],
\end{equation}
which is independently quantized into one of $V=256$ uniform bins. Specifically, we evenly partition the interval $[-1,1]$ and define $257$ bin edges:
\begin{equation}
e_j = -1 + \frac{2j}{256}, \quad j = 0,\ldots,256.
\end{equation}
We use left-closed and right-open bins $[e_j, e_{j+1})$ (with the last bin including the right endpoint), and map $\hat{a}_i$ to a discrete index $q_i \in \{0,\ldots,255\}$, yielding a discrete sequence:
\begin{equation}
\mathbf{q} \in \{0,\ldots,255\}^{M},
\end{equation}
where $M = T \times D$ ($T$ is the number of time frames and $D$ is the action dimension). The sequence can be reshaped into a $T \times D$ 2D temporal--spatial structure to support subsequent masking-based modeling. During decoding, we map token $q$ back to the corresponding bin center to recover continuous actions. We do not adopt the widely used VQ-VAE-based discretization, as VQ-VAE typically introduces strong compression over action sequences, which can overly compress temporal dependencies and inter-dimensional coupling, leading to the loss of fine-grained spatio-temporal information critical for action generation.

\subsection{Temporal--spatial 2D Token Masking} 
\label{sec:mask}
After action quantization, we need to formulate a masking strategy. Since robotic actions possess a two-dimensional structure, applying a 1D masking strategy uniformly to all tokens is suboptimal, as it weakens the model's ability to capture temporal dependencies. To address this issue, we developed a temporal--spatial two-dimensional masking strategy, as illustrated in Fig.~\ref{fig:temporal_spatial_masking}. We reshape the discretized action token sequence into:
\begin{equation}
\mathbf{A} \in \{0, \dots, V-1\}^{T \times D},
\end{equation}
where $\mathbf{A}$ denotes the discretized 2D action token map. For each sample, we draw:
\begin{equation}
t \sim \mathrm{Uniform}(0,1),
\end{equation}
and compute the masking ratio via a cosine schedule:
\begin{equation}
r = \cos\left(\frac{\pi}{2} t\right),
\end{equation}
We apply a two-stage masking strategy. 
In the temporal masking stage, we randomly select $\mathrm{round}(rT)$ time frames
and mask all $D$ action dimensions within each selected frame. 
All tokens within the selected time steps are simultaneously marked as masked, encouraging the model to capture inter-temporal dependencies. 
In the spatial masking stage, for each frame not fully covered by temporal masking, we randomly select $\mathrm{round}(r \times D)$ tokens for masking, introducing finer-grained structural perturbations and enhancing intra-action modeling. 
The loss is computed over the union of masked positions from both stages. 
After temporal--spatial masking, we obtain a masked 2D token map for subsequent prediction training.

\begin{table*}[t]
\caption{Performance comparison on LIBERO. The performance metric is success rate (\%). Spatial, Object, Goal, and Long are labels for different task categories. \textbf{Bold} indicates the best performance, \underline{underlined} numbers indicates the second-best performance, $\dagger$ denotes non-VLM-based methods, and * indicates that this experiment was reproduced under the same setting as our proposed model, ``Params'' denotes the backbone scale, measured in billions (B).}
\label{libero_result}
\centering
\small
\setlength{\tabcolsep}{6pt}
\renewcommand{\arraystretch}{0.9}
\begin{tabular*}{\textwidth}{@{\extracolsep{\fill}} l l c c c c c c@{}}
\toprule
\multicolumn{2}{c}{\textbf{Model}} & \textbf{Params}
& \textbf{Spatial} & \textbf{Object} & \textbf{Goal} & \textbf{Long} & \textbf{Avg.} \\
\midrule

\multirow{4}{*}{\textit{Large}} &
FlowVLA   & 8.5 & 93.2 & 95.0 & 91.6 & 72.6 & 88.1 \\
& OpenVLA   & 7   & 84.7 & 88.4 & 79.2 & 53.7 & 76.5 \\
& CoT-VLA   & 7   & 87.5 & 91.6 & 87.6 & 69.0 & 81.1 \\
& ThinkAct   & 7   & 88.3 & 91.4 & 87.1 & 70.9 & 84.4 \\
\midrule

\multirow{4}{*}{\textit{Small}}
& $\pi_0$     & 3   & \textbf{96.8} & 98.8 & \underline{95.8} & 85.2 & \underline{94.2} \\
& $\pi_0$-FAST  & 3   & \underline{96.4} & 96.8 & 88.6 & 60.2 & 85.5 \\
& SmolVLA    & 2.2 & 93.0 & 94.0 & 91.0 & 77.0 & 88.8 \\
& GR00T N1    & 2   & 94.4 & 97.6 & 93.0 & \underline{90.6} & 93.9 \\
\midrule

\multirow{5}{*}{\textit{Tiny}} &
Seer$^{\dagger}$(Scratch)  & 0.57 & -- & -- & -- & 78.7 & 78.7 \\
& VLA-OS  & 0.5  & 87.0 & 96.5 & 92.7 & 66.0 & 85.6 \\
& Diffusion Policy$^{\dagger}$  & -- & 78.3 & 92.5 & 68.3 & 50.5 & 72.4 \\
& VLA-Adapter-Pro*    & 0.5 & 95.0 & \underline{99.0} & 94.0 & 80.8 & 92.2 \\
& \textbf{TS-Mask VLA} & \textbf{0.5} & 95.4 & \textbf{99.4} & \textbf{96.2} & \textbf{91.6} & \textbf{95.7}\\
\bottomrule
\end{tabular*}

\end{table*}

\begin{table*}[t]
\caption{Performance comparison on CALVIN ABC$\rightarrow$D. The performance metric is success rate (\%). \textbf{Bold} indicates the best performance, \underline{underlined} numbers indicates the second-best performance, $\dagger$ denotes non-VLM-based methods, ``Params'' denotes the backbone scale, measured in billions (B).}
\centering
\small
\setlength{\tabcolsep}{5pt}
\renewcommand{\arraystretch}{0.9}

\begin{tabular*}{\textwidth}{@{\extracolsep{\fill}}l l c c c c c c c@{}}
\toprule
\multicolumn{2}{c}{\textbf{Model}} & \textbf{Params} & \multicolumn{5}{c}{\textbf{Task completed in a row $\uparrow$}} & \textbf{Avg. len $\uparrow$} \\
\cmidrule(lr){4-8}
 &  & & 1 & 2 & 3 & 4 & 5 &  \\
\midrule

\multirow{6}{*}{\textit{Large}}
& UniVLA & 7 & 95.5 & 85.8 & 75.4 & 66.9 & 56.5 & 3.80 \\

& OpenVLA 
& 7 & 91.3 & 77.8 & 62.0 & 52.1 & 43.5 & 3.27 \\

& OpenVLA-OFT  
& 7 & 96.3 & 89.1 & 82.4 & \underline{75.8} & \underline{66.5} & \underline{4.10} \\



& RoboDual  
& 7 & 94.4 & 82.7 & 72.1 & 62.4 & 54.4 & 3.66 \\

& OpenHelix  
& 7 & \underline{97.1} & \underline{91.4} & \underline{82.8} & 72.6 & 64.1 & 4.08 \\

& ReconVLA 
& 7 & 95.6 & 87.6 & 76.9 & 69.3 & 64.1 & 3.95 \\

\midrule

\multirow{3}{*}{\textit{Small}}
& DeeR 
& 3 & 86.2 & 70.1 & 51.8 & 41.5 & 30.4 & 2.82 \\

& RoboFlamingo
& 3 & 82.4 & 61.9 & 46.6 & 33.1 & 23.5 & 2.48 \\


& SuSIE 
& 1.3 & 87.0 & 69.0 & 49.0 & 38.0 & 26.0 & 2.69 \\

\midrule

\multirow{3}{*}{\textit{Tiny}}

& MoDE$^{\dagger}$ 
& 0.44 & 96.2 & 88.9 & 81.1 & 71.8 & 63.5 & 4.01 \\

& Seer$^{\dagger}$  
& 0.32 & 94.4 & 87.2 & 79.9 & 72.2 & 64.3 & 3.98 \\


& \textbf{TS-Mask VLA}
& \textbf{0.5} & \textbf{97.4} & \textbf{92.5} & \textbf{85.2} & \textbf{77.0} & \textbf{66.9}  & \textbf{4.19} \\

\bottomrule
\end{tabular*}

\label{tab:calvin}
\end{table*}

\subsection{Training and Inference}
\label{sec:Training}
\textbf{Training}\quad
The primary training objective is to predict the masked tokens under conditional denoising. 
For the masked position set $\mathcal{M}$, we employ a cross-entropy loss defined as:
\begin{equation}
\mathcal{L}_{\text{mask}}
=
\sum_{i \in \mathcal{M}}
-\log
p_\theta
\big(
a_i \mid
\tilde{\mathbf{a}}_{\setminus \mathcal{M}},
\mathbf{H}
\big),
\end{equation}
where $\tilde{\mathbf{a}}_{\setminus \mathcal{M}}$ denotes the corrupted input sequence. 
The model predicts the ground-truth token $a_i$ at masked positions conditioned on the multi-layer features $\mathbf{H}$. 
The loss is computed only over masked positions, enabling masked denoising modeling.

To mitigate the discrepancy between training and inference, we further introduce a step unroll loss $\mathcal{L}_{\text{unroll}}$. 
After the first forward pass, high-confidence predictions are filled back into the sequence, and a second prediction is performed on the remaining masked positions. 
$\mathcal{L}_{\text{unroll}}$ defined as the cross-entropy loss at the remaining mask positions during the second forward pass, its calculation method is identical to $\mathcal{L}_{\text{mask}}$  except that the conditional input becomes the backfilled sequence.
The final training objective is a weighted combination:

\begin{equation}
\mathcal{L}
=
\frac{
\mathcal{L}_{\text{mask}}
+
\lambda \mathcal{L}_{\text{unroll}}
}{
1 + \lambda
},
\end{equation}
where $\lambda$ controls the contribution of the unroll loss.
\begin{figure}[t]
    \centering
    \includegraphics[width=\columnwidth]{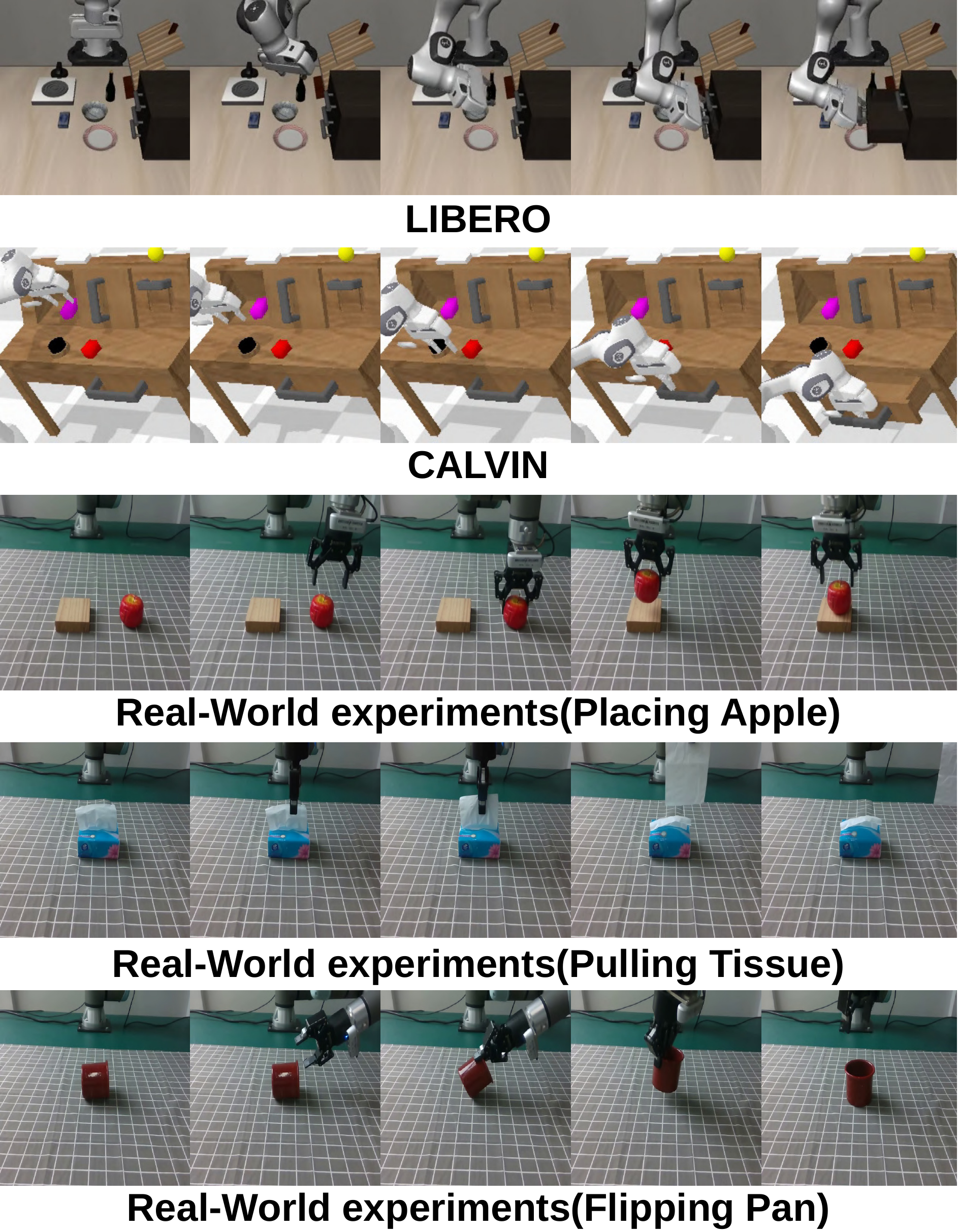}
    \caption{\textbf{All Evaluation environments}, We conduct comprehensive evaluations of TS-Mask VLA in both simulation and real-world settings. In simulation, evaluations are performed on the LIBERO benchmark and CALVIN benchmark. In the real-world experiment, we set up three experimental setups. figure shows our physical experiment.}
    \label{fig:experiment_img}
\end{figure}

\textbf{Inference}\quad
As illustrated in Fig.~\ref{fig:inference}, inference starts from a fully masked action-token sequence $t_0$, where all positions are initialized as \texttt{[MASK]}. The model then performs iterative refinement for $I$ iterations. At iteration $i$, it predicts token probabilities at the currently masked positions and forms a categorical distribution over the $V$ token candidates, from which token values are sampled. We compute a confidence score for each predicted token and, before the next iteration, re-mask the lowest-confidence $m_i=\left\lceil \gamma\!\left(\tfrac{i}{I}\right)\cdot|\Omega_i|\right\rceil$ tokens, where $|\Omega_i|$ denotes the number of tokens filled at iteration $i$ and $\gamma(\cdot)$ is a cosine scheduler that controls the re-masking ratio. As $i$ increases, $\gamma(\tfrac{i}{I})$ decreases, progressively fixing high-confidence predictions and enabling a coarse-to-fine refinement process. This procedure repeats until $i=I$, yielding a fully determined discrete token sequence, which is finally decoded by the action detokenizer into a continuous action sequence.

\section{EXPERIMENTS}

\subsection{Experiment Setup}
We evaluate our method on two widely-used simulated manipulation benchmarks: LIBERO \cite{liu2023libero} and CALVIN\cite{mees2022calvin}. All evaluation environments as shown in Fig.~\ref{fig:experiment_img}. 
For LIBERO, we conduct experiments on four representative task suites, including LIBERO-Object, LIBERO-Goal, LIBERO-Spatial, and LIBERO-Long, which cover object-centric understanding, goal-conditioned generalization, spatial reasoning, and long-horizon manipulation. 
For CALVIN, we adopt the standard ABC-D evaluation protocol to assess the agent's ability to execute multi-step language-conditioned task sequences in a long-horizon setting.
We train TS-Mask VLA on a single NVIDIA RTX 4090 GPU. Following common practice, we fine-tune the VLM backbone with LoRA for parameter-efficient adaptation. Unless otherwise specified, we set the action chunk length to 8.

\subsection{Baselines}
We select recently released and high-performing VLA methods as our baselines, classify them into three categories based on scale, including:
\textbf{Large-scale models:}
FlowVLA\cite{zhong2025flowvla}, OpenVLA\cite{kim2024openvla}, CoT-VLA\cite{zhao2025cot}, ThinkAct\cite{huang2025thinkact}, UniVLA\cite{bu2025univla}, OpenVLA-OFT\cite{OpenVLAOFT}, RoboDual\cite{RoboDual}, OpenHelix\cite{cui2025openhelix}  and ReconVLA\cite{song2025reconvla}.
\textbf{Small-scale models:}
 $\pi_{0}$\cite{black2024pi_0}, $\pi_{0}$-FAST\cite{pertsch2025fast}, SmolVLA\cite{shukor2025smolvla}, GR00T~N1\cite{bjorck2025gr00t}, DeeR\cite{yue2024deer}, RoboFlamingo\cite{li2023vision} and SuSIE\cite{SuSIE}.
\textbf{Tiny-scale models:}
Seer$^{\dagger}$\cite{tian2024predictive}, VLA-OS\cite{gao2025vla}, Diffusion Policy$^{\dagger}$\cite{chi2025diffusion}, VLA-Adapter\cite{Vla-adapter} and MoDE\cite{MODE}.

\subsection{Results and Analysis.}

\textbf{LIBERO results.}
As shown in Table~\ref{libero_result}, our method achieves an average success rate of 95.7\% on the LIBERO benchmark, demonstrating stable and consistent performance across all task suites. Although we employ a lightweight 0.5B backbone, our model surpasses FlowVLA—whose parameter scale is 19$\times$ larger—by 7.6\% in average success rate. In addition, our approach outperforms the strong baseline $\pi_0$ by 1.5\%. The advantages of our method are particularly pronounced in long-horizon tasks, where it exceeds the classical model GR00T~N1 by 1\%. These improvements can be attributed to the effective modeling of spatiotemporal relationships among actions.
\begin{figure*}[t]
    \centering
    \includegraphics[width=\textwidth]{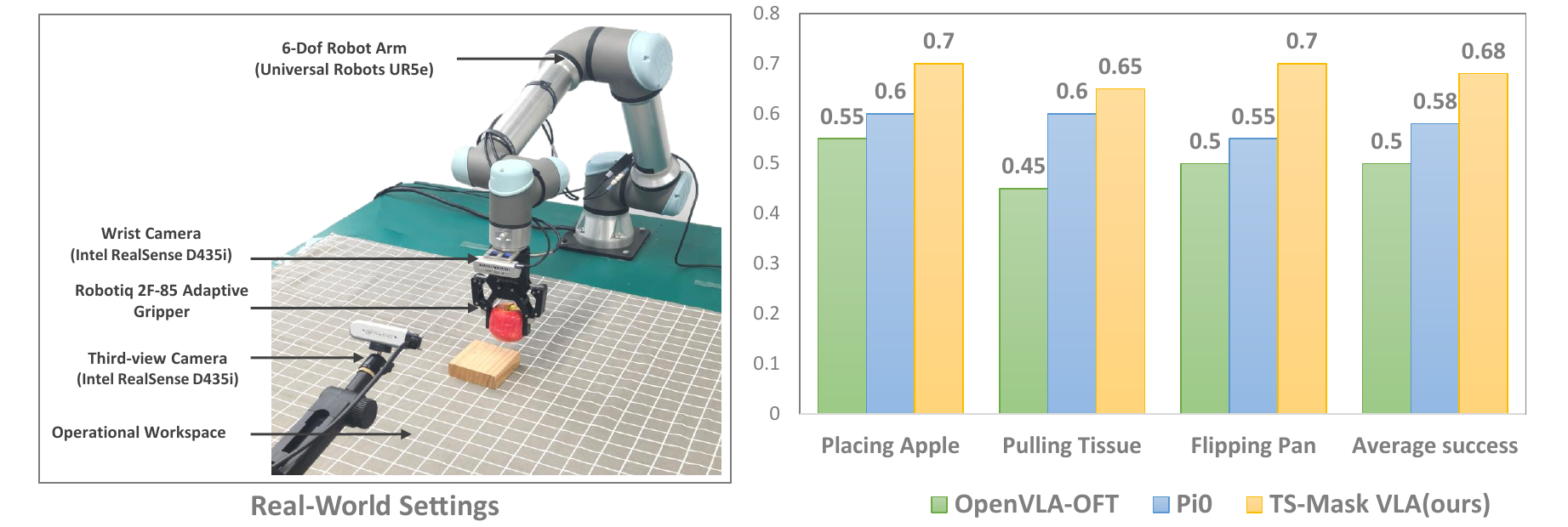}
    \caption{\textbf{Real-World Settings and results.}
    }
    \label{fig:real_world}
\end{figure*}

\textbf{CALVIN results.}
As shown in Table~\ref{tab:calvin}, our method achieves the best overall performance on the CALVIN benchmark. 
With only a 0.5B backbone, TS-Mask VLA attains the highest average sequence length of 4.19, outperforming all 7B-scale models.
Compared with OpenVLA-OFT, our method achieves a higher 5-task completion rate (66.9\% vs. 66.5\%) and a better average length (4.19 vs. 4.10) while using 14$\times$ fewer parameters. 
More notably, compared with OpenVLA, we significantly improve long-horizon performance, increasing the 4-task completion rate from 52.1\% to 77.0\% and the 5-task completion rate from 43.5\% to 66.9\%.
These results demonstrate the strong parameter efficiency of our approach and its effectiveness in modeling long-horizon temporal dependencies.

\subsection{Ablation Study}

\textbf{Effectiveness of Temporal-Spatial Masking Strategy.}
To evaluate the impact of mask structures on discrete action generation, we conducted ablation experiments comparing one-dimensional (1D) masks with two-dimensional (2D) temporal--spatial masking strategies. 

As shown in Table~\ref{tab:mask_ablation}, the 2D masking approach consistently outperformed the 1D masking baseline across all LIBERO suites. It yields approximately 1\% gains on the Spatial, Object, and Goal datasets, while achieving a significant 6.6\% improvement in success rate on the Long dataset, which relies more heavily on temporal reasoning. These results demonstrate that explicitly modeling temporal--spatial structures provides stronger temporal--spatial coupling for discrete action generation, with particularly pronounced effects in long-sequence task scenarios.


{\setlength{\tabcolsep}{7pt} 
\begin{table}[t]
\centering
\caption{Ablation on Masking Strategy (Success Rate \%),\textbf{Bold} indicates the best performance.}
\begin{tabular}{lcccc}
\toprule
Mask Strategy & Spatial & Object & Goal & Long \\
\midrule
1D & 94.6 & 98.4 & 95.2 & 85.0 \\
2D & \textbf{95.4} & \textbf{99.4} & \textbf{96.2} & \textbf{91.6} \\
\bottomrule
\end{tabular}
\label{tab:mask_ablation}
\end{table}
}





\begin{table}[t]
\centering
\caption{Ablation on Step Unroll Strength (LIBERO-Spatial Success Rate \%), \textbf{Bold} indicates the best performance.}
\label{tab:unroll_ablation}

\renewcommand{\arraystretch}{1.1}
\setlength{\tabcolsep}{7pt} 

\begin{tabular}{lccc}
\toprule
 & No Unroll & $\lambda=0.5$ & $\lambda=1.0$ \\
\midrule
LIBERO-Spatial & 90.8 & \textbf{95.4} & 91.5 \\
\bottomrule
\end{tabular}
\end{table}

\textbf{Effectiveness of Step Unroll Strength.}
To investigate the effect of the step unroll loss, we conduct an ablation study with different unroll strengths $\lambda$, including no unroll, $\lambda = 0.5$, and $\lambda = 1.0$.

As reported in Table~\ref{tab:unroll_ablation}, introducing a moderate unroll strength ($\lambda = 0.5$) significantly improves performance, increasing the LIBERO-Spatial success rate from 90.8\% to 95.4\%, a gain of 4.6\%. In contrast, an overly strong unroll constraint ($\lambda = 1.0$) reduces the success rate to 91.5\%, suggesting that excessive regularization may hinder the learning of the primary masking objective. 

These results indicate that while step unroll effectively mitigates train-test discrepancy, its contribution must be carefully balanced with the main masking loss to avoid over-regularization.

\subsection{Real-World Experiment}
\textbf{Setup and Tasks.}
Our real-world experimental setup consists of a 6-DoF UR5e robotic arm equipped with a ROBOTIQ-85 gripper. We deploy two Intel RealSense D435i depth cameras to provide RGB observations of the manipulated objects: one is mounted statically at the front of the workspace to capture the entire operational space, while the other is wrist-mounted to offer close-range observations of the manipulated objects. We collect three categories of tasks, namely placing an apple on the table, pulling out a tissue, and flipping a pan. For each task category, we place the objects at different positions to increase the task difficulty.

\textbf{Results and Analysis.}
We evaluate each method with 20 trials per task. As shown in Fig.~\ref{fig:real_world}, our approach consistently outperforms the two baselines, OpenVLA-OFT and $\pi_{0}$, across all tasks. 
Our method demonstrates stronger spatial understanding and generalization in diverse real-world scenarios, which we attribute to the proposed discrete diffusion action expert. Specifically, it effectively receives hierarchical feature injections from multiple intermediate layers of the VLM, enabling the model to leverage both low-level geometric and boundary cues as well as high-level semantic information and task intent. This hierarchical conditioning facilitates the learning of more robust cross-scene representations.
In addition, the introduced temporal--spatial masking mechanism further strengthens the encoding of action-relevant regions and key time steps within the action sequence. As a result, even under target position shifts and complex background variations, our model can generate consistent and accurate action sequences.
\section{Conclusion}
We propose TS-Mask VLA, a 2D Temporal--Spatial Masking approach for Vision-Language-Action model with effective Bridging. Our approach effectively injects multimodal knowledge into an discrete diffusion action expert and employs a spatio-temporal 2D masking strategy to strengthen temporal dependencies and cross-dimensional coupling in action sequences. The resulting discrete diffusion action expert enables accurate and consistent action generation. Despite using a tiny backbone, our method demonstrates strong effectiveness on both simulation benchmarks and real-world robotic tasks.

\textbf{Limitations and Future Work} 
We have only discussed the experimental results of TS-Mask VLA under resource-constrained scenarios; in future work, we will investigate its performance under abundant training resources.













\begin{thebibliography}{10}
\providecommand{\url}[1]{#1}
\csname url@samestyle\endcsname
\providecommand{\newblock}{\relax}
\providecommand{\bibinfo}[2]{#2}
\providecommand{\BIBentrySTDinterwordspacing}{\spaceskip=0pt\relax}
\providecommand{\BIBentryALTinterwordstretchfactor}{4}
\providecommand{\BIBentryALTinterwordspacing}{\spaceskip=\fontdimen2\font plus
\BIBentryALTinterwordstretchfactor\fontdimen3\font minus
  \fontdimen4\font\relax}
\providecommand{\BIBforeignlanguage}[2]{{%
\expandafter\ifx\csname l@#1\endcsname\relax
\typeout{** WARNING: IEEEtran.bst: No hyphenation pattern has been}%
\typeout{** loaded for the language `#1'. Using the pattern for}%
\typeout{** the default language instead.}%
\else
\language=\csname l@#1\endcsname
\fi
#2}}
\providecommand{\BIBdecl}{\relax}
\BIBdecl

\bibitem{liang2025discrete}
Z.~Liang, Y.~Li, T.~Yang, C.~Wu, S.~Mao, T.~Nian, L.~Pei, S.~Zhou, X.~Yang,
  J.~Pang \emph{et~al.}, ``Discrete diffusion vla: Bringing discrete diffusion
  to action decoding in vision-language-action policies,'' \emph{arXiv preprint
  arXiv:2508.20072}, 2025.

\bibitem{wen2025llada}
Y.~Wen, H.~Li, K.~Gu, Y.~Zhao, T.~Wang, and X.~Sun, ``Llada-vla: Vision
  language diffusion action models,'' \emph{arXiv preprint arXiv:2509.06932},
  2025.

\bibitem{kim2024openvla}
M.~J. Kim, K.~Pertsch, S.~Karamcheti, T.~Xiao, A.~Balakrishna, S.~Nair,
  R.~Rafailov, E.~Foster, G.~Lam, P.~Sanketi \emph{et~al.}, ``Openvla: An
  open-source vision-language-action model,'' \emph{arXiv preprint
  arXiv:2406.09246}, 2024.

\bibitem{chi2025diffusion}
C.~Chi, Z.~Xu, S.~Feng, E.~Cousineau, Y.~Du, B.~Burchfiel, R.~Tedrake, and
  S.~Song, ``Diffusion policy: Visuomotor policy learning via action
  diffusion,'' \emph{The International Journal of Robotics Research}, vol.~44,
  no. 10-11, pp. 1684--1704, 2025.

\bibitem{shukor2025smolvla}
M.~Shukor, D.~Aubakirova, F.~Capuano, P.~Kooijmans, S.~Palma, A.~Zouitine,
  M.~Aractingi, C.~Pascal, M.~Russi, A.~Marafioti \emph{et~al.}, ``Smolvla: A
  vision-language-action model for affordable and efficient robotics,''
  \emph{arXiv preprint arXiv:2506.01844}, 2025.

\bibitem{reuss2024multimodal}
M.~Reuss, {\"O}.~E. Ya{\u{g}}murlu, F.~Wenzel, and R.~Lioutikov, ``Multimodal
  diffusion transformer: Learning versatile behavior from multimodal goals,''
  \emph{arXiv preprint arXiv:2407.05996}, 2024.

\bibitem{Vla-adapter}
Y.~Wang, P.~Ding, L.~Li, C.~Cui, Z.~Ge, X.~Tong, W.~Song, H.~Zhao, W.~Zhao,
  P.~Hou \emph{et~al.}, ``Vla-adapter: An effective paradigm for tiny-scale
  vision-language-action model,'' \emph{arXiv preprint arXiv:2509.09372}, 2025.

\bibitem{javed2024intermask}
M.~G. Javed, C.~Guo, L.~Cheng, and X.~Li, ``Intermask: 3d human interaction
  generation via collaborative masked modeling,'' \emph{arXiv preprint
  arXiv:2410.10010}, 2024.

\bibitem{liu2023libero}
B.~Liu, Y.~Zhu, C.~Gao, Y.~Feng, Q.~Liu, Y.~Zhu, and P.~Stone, ``Libero:
  Benchmarking knowledge transfer for lifelong robot learning,'' \emph{Advances
  in Neural Information Processing Systems}, vol.~36, pp. 44\,776--44\,791,
  2023.

\bibitem{mees2022calvin}
O.~Mees, L.~Hermann, E.~Rosete-Beas, and W.~Burgard, ``Calvin: A benchmark for
  language-conditioned policy learning for long-horizon robot manipulation
  tasks,'' \emph{IEEE Robotics and Automation Letters}, vol.~7, no.~3, pp.
  7327--7334, 2022.

\bibitem{black2024pi_0}
K.~Black, N.~Brown, D.~Driess, A.~Esmail, M.~Equi, C.~Finn, N.~Fusai, L.~Groom,
  K.~Hausman, B.~Ichter \emph{et~al.}, ``{$\pi_0$}: A vision-language-action
  flow model for general robot control,'' \emph{arXiv preprint
  arXiv:2410.24164}, 2024.

\bibitem{Rt-1}
A.~Brohan, N.~Brown, J.~Carbajal, Y.~Chebotar, J.~Dabis, C.~Finn,
  K.~Gopalakrishnan, K.~Hausman, A.~Herzog, J.~Hsu \emph{et~al.}, ``Rt-1:
  Robotics transformer for real-world control at scale,'' \emph{arXiv preprint
  arXiv:2212.06817}, 2022.

\bibitem{Rt-2}
B.~Zitkovich, T.~Yu, S.~Xu, P.~Xu, T.~Xiao, F.~Xia, J.~Wu, P.~Wohlhart,
  S.~Welker, A.~Wahid \emph{et~al.}, ``Rt-2: Vision-language-action models
  transfer web knowledge to robotic control,'' in \emph{Conference on Robot
  Learning}.\hskip 1em plus 0.5em minus 0.4em\relax PMLR, 2023, pp. 2165--2183.

\bibitem{austin2021structured}
J.~Austin, D.~D. Johnson, J.~Ho, D.~Tarlow, and R.~Van Den~Berg, ``Structured
  denoising diffusion models in discrete state-spaces,'' \emph{Advances in
  neural information processing systems}, vol.~34, pp. 17\,981--17\,993, 2021.

\bibitem{hoogeboom2021argmax}
E.~Hoogeboom, D.~Nielsen, P.~Jaini, P.~Forr{\'e}, and M.~Welling, ``Argmax
  flows and multinomial diffusion: Learning categorical distributions,''
  \emph{Advances in neural information processing systems}, vol.~34, pp.
  12\,454--12\,465, 2021.

\bibitem{gu2022vector}
S.~Gu, D.~Chen, J.~Bao, F.~Wen, B.~Zhang, D.~Chen, L.~Yuan, and B.~Guo,
  ``Vector quantized diffusion model for text-to-image synthesis,'' in
  \emph{Proceedings of the IEEE/CVF conference on computer vision and pattern
  recognition}, 2022, pp. 10\,696--10\,706.

\bibitem{li2022diffusion}
X.~Li, J.~Thickstun, I.~Gulrajani, P.~S. Liang, and T.~B. Hashimoto,
  ``Diffusion-lm improves controllable text generation,'' \emph{Advances in
  neural information processing systems}, vol.~35, pp. 4328--4343, 2022.

\bibitem{chang2022maskgit}
H.~Chang, H.~Zhang, L.~Jiang, C.~Liu, and W.~T. Freeman, ``Maskgit: Masked
  generative image transformer,'' in \emph{Proceedings of the IEEE/CVF
  conference on computer vision and pattern recognition}, 2022, pp.
  11\,315--11\,325.

\bibitem{nie2025large}
S.~Nie, F.~Zhu, Z.~You, X.~Zhang, J.~Ou, J.~Hu, J.~Zhou, Y.~Lin, J.-R. Wen, and
  C.~Li, ``Large language diffusion models,'' \emph{arXiv preprint
  arXiv:2502.09992}, 2025.

\bibitem{oquab2023dinov2}
M.~Oquab, T.~Darcet, T.~Moutakanni, H.~Vo, M.~Szafraniec, V.~Khalidov,
  P.~Fernandez, D.~Haziza, F.~Massa, A.~El-Nouby \emph{et~al.}, ``Dinov2:
  Learning robust visual features without supervision,'' \emph{arXiv preprint
  arXiv:2304.07193}, 2023.

\bibitem{zhai2023sigmoid}
X.~Zhai, B.~Mustafa, A.~Kolesnikov, and L.~Beyer, ``Sigmoid loss for language
  image pre-training,'' in \emph{Proceedings of the IEEE/CVF international
  conference on computer vision}, 2023, pp. 11\,975--11\,986.

\bibitem{hui2024qwen2}
B.~Hui, J.~Yang, Z.~Cui, J.~Yang, D.~Liu, L.~Zhang, T.~Liu, J.~Zhang, B.~Yu,
  K.~Lu \emph{et~al.}, ``Qwen2.5-coder technical report,'' \emph{arXiv preprint
  arXiv:2409.12186}, 2024.

\bibitem{zhong2025flowvla}
Z.~Zhong, H.~Yan, J.~Li, X.~Liu, X.~Gong, W.~Song, J.~Chen, and H.~Li,
  ``Flowvla: Thinking in motion with a visual chain of thought,'' \emph{arXiv
  e-prints}, pp. arXiv--2508, 2025.

\bibitem{zhao2025cot}
Q.~Zhao, Y.~Lu, M.~J. Kim, Z.~Fu, Z.~Zhang, Y.~Wu, Z.~Li, Q.~Ma, S.~Han,
  C.~Finn \emph{et~al.}, ``Cot-vla: Visual chain-of-thought reasoning for
  vision-language-action models,'' in \emph{Proceedings of the Computer Vision
  and Pattern Recognition Conference}, 2025, pp. 1702--1713.

\bibitem{huang2025thinkact}
C.-P. Huang, Y.-H. Wu, M.-H. Chen, Y.-C.~F. Wang, and F.-E. Yang, ``Thinkact:
  Vision-language-action reasoning via reinforced visual latent planning,''
  \emph{arXiv preprint arXiv:2507.16815}, 2025.

\bibitem{bu2025univla}
Q.~Bu, Y.~Yang, J.~Cai, S.~Gao, G.~Ren, M.~Yao, P.~Luo, and H.~Li, ``Univla:
  Learning to act anywhere with task-centric latent actions,'' \emph{arXiv
  preprint arXiv:2505.06111}, 2025.

\bibitem{OpenVLAOFT}
M.~J. Kim, C.~Finn, and P.~Liang, ``Fine-tuning vision-language-action models:
  Optimizing speed and success,'' \emph{arXiv preprint arXiv:2502.19645}, 2025.

\bibitem{RoboDual}
Q.~Bu, H.~Li, L.~Chen, J.~Cai, J.~Zeng, H.~Cui, M.~Yao, and Y.~Qiao, ``Towards
  synergistic, generalized, and efficient dual-system for robotic
  manipulation,'' \emph{arXiv preprint arXiv:2410.08001}, 2024.

\bibitem{cui2025openhelix}
C.~Cui, P.~Ding, W.~Song, S.~Bai, X.~Tong, Z.~Ge, R.~Suo, W.~Zhou, Y.~Liu,
  B.~Jia \emph{et~al.}, ``Openhelix: A short survey, empirical analysis, and
  open-source dual-system vla model for robotic manipulation,'' \emph{arXiv
  preprint arXiv:2505.03912}, 2025.

\bibitem{song2025reconvla}
W.~Song, Z.~Zhou, H.~Zhao, J.~Chen, P.~Ding, H.~Yan, Y.~Huang, F.~Tang,
  D.~Wang, and H.~Li, ``Reconvla: Reconstructive vision-language-action model
  as effective robot perceiver,'' \emph{arXiv preprint arXiv:2508.10333}, 2025.

\bibitem{pertsch2025fast}
K.~Pertsch, K.~Stachowicz, B.~Ichter, D.~Driess, S.~Nair, Q.~Vuong, O.~Mees,
  C.~Finn, and S.~Levine, ``Fast: Efficient action tokenization for
  vision-language-action models,'' \emph{arXiv preprint arXiv:2501.09747},
  2025.

\bibitem{bjorck2025gr00t}
J.~Bjorck, F.~Casta{\~n}eda, N.~Cherniadev, X.~Da, R.~Ding, L.~Fan, Y.~Fang,
  D.~Fox, F.~Hu, S.~Huang \emph{et~al.}, ``Gr00t n1: An open foundation model
  for generalist humanoid robots,'' \emph{arXiv preprint arXiv:2503.14734},
  2025.

\bibitem{yue2024deer}
Y.~Yue, Y.~Wang, B.~Kang, Y.~Han, S.~Wang, S.~Song, J.~Feng, and G.~Huang,
  ``Deer-vla: Dynamic inference of multimodal large language models for
  efficient robot execution,'' \emph{Advances in Neural Information Processing
  Systems}, vol.~37, pp. 56\,619--56\,643, 2024.

\bibitem{li2023vision}
X.~Li, M.~Liu, H.~Zhang, C.~Yu, J.~Xu, H.~Wu, C.~Cheang, Y.~Jing, W.~Zhang,
  H.~Liu \emph{et~al.}, ``Vision-language foundation models as effective robot
  imitators,'' \emph{arXiv preprint arXiv:2311.01378}, 2023.

\bibitem{SuSIE}
K.~Black, M.~Nakamoto, P.~Atreya, H.~Walke, C.~Finn, A.~Kumar, and S.~Levine,
  ``Zero-shot robotic manipulation with pretrained image-editing diffusion
  models,'' \emph{arXiv preprint arXiv:2310.10639}, 2023.

\bibitem{tian2024predictive}
Y.~Tian, S.~Yang, J.~Zeng, P.~Wang, D.~Lin, H.~Dong, and J.~Pang, ``Predictive
  inverse dynamics models are scalable learners for robotic manipulation,''
  \emph{arXiv preprint arXiv:2412.15109}, 2024.

\bibitem{gao2025vla}
C.~Gao, Z.~Liu, Z.~Chi, J.~Huang, X.~Fei, Y.~Hou, Y.~Zhang, Y.~Lin, Z.~Fang,
  Z.~Jiang \emph{et~al.}, ``Vla-os: Structuring and dissecting planning
  representations and paradigms in vision-language-action models,'' \emph{arXiv
  preprint arXiv:2506.17561}, 2025.

\bibitem{MODE}
M.~Reuss, J.~Pari, P.~Agrawal, and R.~Lioutikov, ``Efficient diffusion
  transformer policies with mixture of expert denoisers for multitask
  learning,'' \emph{arXiv preprint arXiv:2412.12953}, 2024.

\end{thebibliography}
\end{document}